# Multi-Layer Feature Fusion with Cross-Channel Attention-Based U-Net for Kidney Tumor Segmentation

**FNU Neha, Arvind K. Bansal**
Department of Computer Science
Kent State University, Kent, OH 44242, USA
Neha@Kent.edu; akbansal@kent.edu

***Abstract*** - Renal tumors, especially renal cell carcinoma(RCC), show significant heterogeneity, posing challenges for diagnosis using radiology images: MRI, echocardiograms, and CT scans. U-Net based deep learning techniques are emerging as a promising approach for automated medical image segmentation for minimally invasive diagnosis of renal tumors. However, current techniques need further improvements in accuracy to become clinically useful to radiologists. In this study, we present an improved U-Net based model for end-to-end automated semantic segmentation of CT scan images to identify renal tumors. The model uses residual connections across convolution layers, integrates a multi-layer feature fusion (MFF) and cross-channel attention (CCA) within encoder-blocks, and skip-connection augmented with additional information derived using MFF and CCA. We evaluated our model on the KiTS19 dataset using 210 patients' data. For kidney segmentation, our model achieves *Dice Similarity Coefficient* (DSC) of 0.97 and *Jaccard index* (JI) of 0.95. For renal tumor segmentation, our model achieves DSC of 0.96, and JI of 0.91. Based upon the comparison of available DSC scores, our model outperforms the current leading models.

## 1. Introduction

Kidney is a bean-shaped multi-functional vital organ. Its major functions are 1) blood-filtration to eliminate waste, toxins, and excess fluids; 2) selective reabsorption of vital minerals to maintain electrolyte balance; 3) secretion of many essential hormones; 4) control of fluid volume to regulate blood-pressure; 5) maintenance of acid-base equilibrium of blood 6) facilitate metabolism of vitamin D by converting it into an active form [1].

Renal cancer (cancer related to kidney) is a major cause of renal failure, resulting in morbidity, costly long-term maintenance, and mortality [2 - 4]. Globally, around 400,000 cases of renal cancer occur every year, resulting in around 175,000 deaths [4]. Renal cancer's etiology involves several risk factors [1, 5]. The risk factors include (1) genetics; (2) lifestyle choices: tobacco smoking, obesity, alcohol intake, sedentary lifestyle, diet; (3) medical history: hypertension, chronic kidney diseases, kidney stones, and diabetes [3, 5].

Renal cancer is detected incidentally while testing for other diseases or at a late stage due to the lack of visual symptoms, the lack of confirming serological tests, and the lack of pathological symptoms in patients at early stages [1, 3]. At late stages, it becomes possibly malignant, resulting in renal failure and mortality [1, 3].

The treatment for renal cancer includes partial or radical nephrectomy, with serious post-surgical complications, including increased risk of chronic kidney disease (CKD), cancer relapse, cardiovascular problems, and eventually End Stage Kidney Disease (ESKD) [3]. In addition, there is an acute shortage of kidney care workforce, including nephrologists, especially in the developing countries [2]. To address these issues, there is a need for an automated and accurate system to detect renal cancers at an early stage.

Serological protein-biomarkers from blood serums or urine are noninvasive indicators for an early detection of renal cancer [6]. However, current biomarkers suffer from the lack of specificity, have not yet been validated, and cannot localize or give the extent of tumor growth [6]. Another minimally invasive option is radiological images to detect the kidney tumors (both benign and malignant), and localize and estimate the extent of tumor growth [7]. Computed Tomography (CT) scans are the choice for renal cancer diagnosis due to their effectiveness and lower cost compared to Magnetic Resonance Imaging (MRI) and Positron Emission Tomography (PET) [7].

Currently, radiologists manually delineate radiology images, which is a time-consuming and tiring process [8]. It is prone to inconsistencies due to discrepancies between radiologists' reports, cognitive and perceptual errors, and fatigue errors of radiologists [8]. Extensive studies reveal an average error-rate of around 3–5% in manual delineation [7]. At 4% extrapolated error-rate, based upon one billion radiology image processed every year, approximately 40 million radiologist-errors occur every year [8]. Thus, there is a need for an automated accurate segmentation of renal tumors.

Advancements in deep learning techniques, especially variants of U-Net-based models [9], integrated with self-attention [10], channel and spatial attention [11] and residual connections [12], are being experimented for automated medical image analysis for renal cancer detection [13-19]. Previous studies use a combination of (1) convolution layers in encoder-block for feature-extraction; (2) a combination of spatial attention and channel attention for identifying important features and dependencies between image-patches; (3) averaging of feature-maps in channel-wise attention; (4) combining feature-information across encoder-blocks using residual connections. However, automated segmentation of renal tumors remains challenging due to low-intensity contrast with surrounding organs, aggregated by irregular shapes [20].

In this research, we propose an enhanced U-Net variant, augmented with the integration of multilayer feature fusion (MFF), use of residual connections between encoder-blocks, and cross-channel attention (CCA) at the encoder-level of U-Net, and enhanced skip-connection, augmented with MFF and CAA features information, between encoder-blocks and the corresponding decoder-blocks, to improve the accuracy of automated segmentation. MFF fuses features derived from multiple convolution layers using residual links between the layers. CCA provides weights to different convolution windows in an image by averaging different feature-maps for each window. Skip-connection between encoders and the corresponding decoders has significant information improvement due to the information derived using MFF and CCA. The resulting model significantly recovers information lost in encoder-blocks, during encoding, and the corresponding decoder-blocks, during reconstruction of the predicted segment during the reconstruction of images in decoder phase. Comparison of our approach with other leading models shows significant improvement in accuracy of renal tumor segmentation.

Our research contributions are:

(1) Introduction of CCA in the encoder-block which identifies important features;

(2) Integration of residual connections, MFF and CCA to significantly reduce the information-loss during encoding;

(3) Enhancement of information-flow derived by integrating MFF and CCA to the decoder using skip-connection.

The paper is organized as follows. Section 2 describes the background. Section 3 describes related work. Section 4 discusses our approach. Section 5 describes the implementation. Section 6 discusses the results and limitations. Section 7 concludes and discusses future work.

## 2. Background

### 2.1. U-Net

U-Net comprises three major components: (1) down-sampling: extracting feature-maps using a convolution-based stack of encoder-blocks; (2) up-sampling: reconstructing the predicted segment using convolution-transpose and fusing skip-connection based information in a stack of decoder-blocks; (3) skip-connections between each encoder and the corresponding decoder to recover the information-loss during the application of convolution and pooling in the encoder phase, and convolution-transpose in the decoding phase. Each encoder comprises sequence of three layers: (1) convolution filters to extract feature-maps; (2) *Rectified Linear Unit* (RELU) to suppress noise; (3) pooling to compresses feature-maps. Finally, $1 \times 1$ convolutional layer produces the segmentation map, categorizing each pixel of the input image.

### 2.2. Visual Attention

Attention enables neural networks to focus on relevant parts of input data, keeping relevant dependencies between the sequence of entities (image-patches), and dynamically allocating computational resources to enhance performance in tasks such as image recognition [10]. Channel attention prioritizes important features by computing importance weights for individual channels through pooling operation followed by multilayer perceptrons and an activation function [11]. Spatial attention emphasizes specific locations of the features by performing pooling operation along channel axis followed by convolution operation and activation function [11]. Global Average Pooling (GAP) is used in channel attention to summarize each feature-map by averaging all spatial locations, reducing each feature-map to a single scalar value. This step emphasizes the global context of each channel but removes spatial information.

**2.3. Residual Connection**

A residual connection connects the previous encoder(s) (or decoder) block to the next layer [12]. If $x$ is the output from the $(i-1)_{th}$ layer, and $F(x)$ is the output of the $i_{th}$ layer, then input $H(x)$ for the $(i+1)_{th}$ layer is equal to $F(x) + x$. The direct flow of information using residual links facilitates improved learning and reduces vanishing gradient problem [11].

## 3. Related Work

Related leading models use variants of U-net: ensemble of U-Nets or integration of U-Net with variants of attention and/or residual connections [13-18]. An attention U-Net integrates attention within encoder-decoder architecture to enhance segmentation accuracy by selectively focusing on informative regions of the input image [11, 16-18]. Residual U-Net introduces residual connections within encoder-decoder blocks to alleviate the vanishing gradient problem [12].

Causey et al. proposed an ensemble of two U-Nets, designed to predict kidney and tumor segmentation separately [13]. The limitation of this scheme is that ensemble's voting mechanism for error reduction can also amplify shared mistakes or biases rather than correcting them, lowering the accuracy.

Cruz et al. proposed AlexNet based CNN for CT slice classification, followed by U-Net for kidney segmentation and post-processing to reduce false positives [14]. The scheme does not perform tumor segmentation. Although AlexNet partially reduces overfitting problem using dropouts, it lacks residual connections for information-flow. As pointed out by the authors, CT slice classification and false-positive reduction steps do not guarantee quantitative improvement in kidney segmentation, as false-positives can still be included.

Wen et al. proposed an encoder-decoder framework, which integrates residual connections and channel attention [15]. Gohil et al. enhance U-Net model with spatial and channel attention [16]. Guo et al. combined U-Net with residual and attention blocks and used two-stage model to derive tumor segmentation [17]. In the first stage, they used threshold to derive rough segmentation and the Region-of-Interest (ROI). The second stage processed ROIs for tumour-segmentation. The channel attention modules in all these schemes use multilayer perceptron or fully connected layer after the pooling operation, treating the input as a flat vector, disregarding spatial structure. Qian et el. combined SWIN transformer and U-Net to capture long-range dependencies between image segments and local features derived by convolution [18]. The use of SWIN transformer also handles high-resolution images efficiently. However, All these methods do not capture the information-loss due to the limited residual links in the attention module and the lack of correspondence between nearby channels along with their weights, lowering their overall accuracy in segmenting renal tumor (see Section 6, Table 1).

Recently, Yuan et al. integrated a variant of MFF with squeeze-and-excitation attention mechanism to automate the segmentation for liver tumor [19]. They use spatial and channel attention and use global average pooling to assign weights to specific channels. However, their approach does not use residual links, lowering the information-flow to the decoders.

Our model integrates residual connections based multilayer feature-fusion, cross-channel attention, and augmented skip-connection to preserve the multi-level information-flow between encoder-decoder blocks. The deep intertwining of residual links and cross-channel attention significantly improves the segmentation accuracy.

## 4. Our Approach

Our approach, as shown in Fig, 1, uses U-Net architecture as a backbone. Each encoder-block is augmented with residual connections for sharing features between each convolution block, Multilayer Feature Fusion Block (MFF), and Cross Channel Attention block (CCA). Skip-connections employ residual links over MFF + CCA augmented encoder output to fill in the information-gaps between encoder-layers and the corresponding decoder-layers. Within each encoder-block, three convolutional blocks are used, along with the residual links. After batch normalization and element-wise summation with the convolutional block outputs, ReLU activation is applied. Each decoder-block concatenates information derived from the corresponding encoder and reconstructed feature-map.

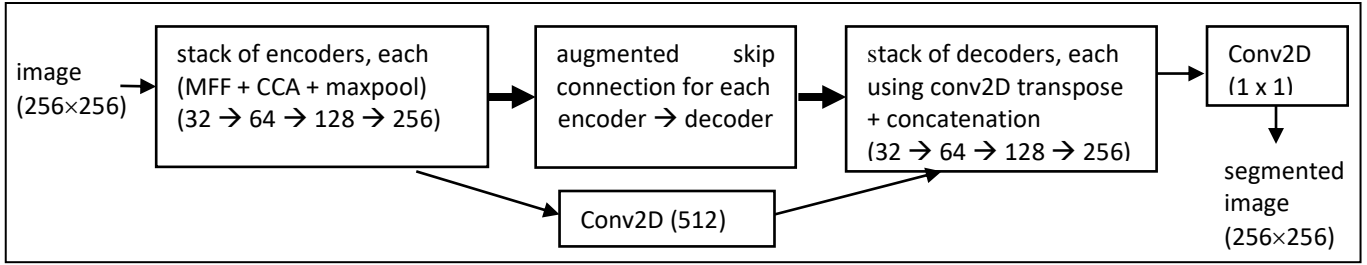

Fig. 1: A schematic of overall architecture

A bridge layer between the last encoder-block and the first decoder-block extracts 512 feature-maps. Spatial information is restored in the decoder-blocks (256, 128, 64, 32) by up-sampling feature-maps, using convolution-transpose. Features transferred through augmented skip-connections are concatenated with the up-sampled features. Finally, a 1×1 convolution and the SoftMax function are used to generate the segmented mask, delineating target structures.

### 4.1 Encoder Block

Encoder-block is a stack of encoders (32, 64, 128, 256). Each encoder comprises MFF layer, CCA layer and 2×2 max-pooling layer for down-sampling. The output of a MFF layer is passed to the corresponding CCA layer. The output of a CCA layer is fed to the corresponding max-pooling layer and the augmented skip-connection. The output of a max-pooling layer is fed to the following encoder-block. The overall schematic of encoder-block components is illustrated in Fig. 2.

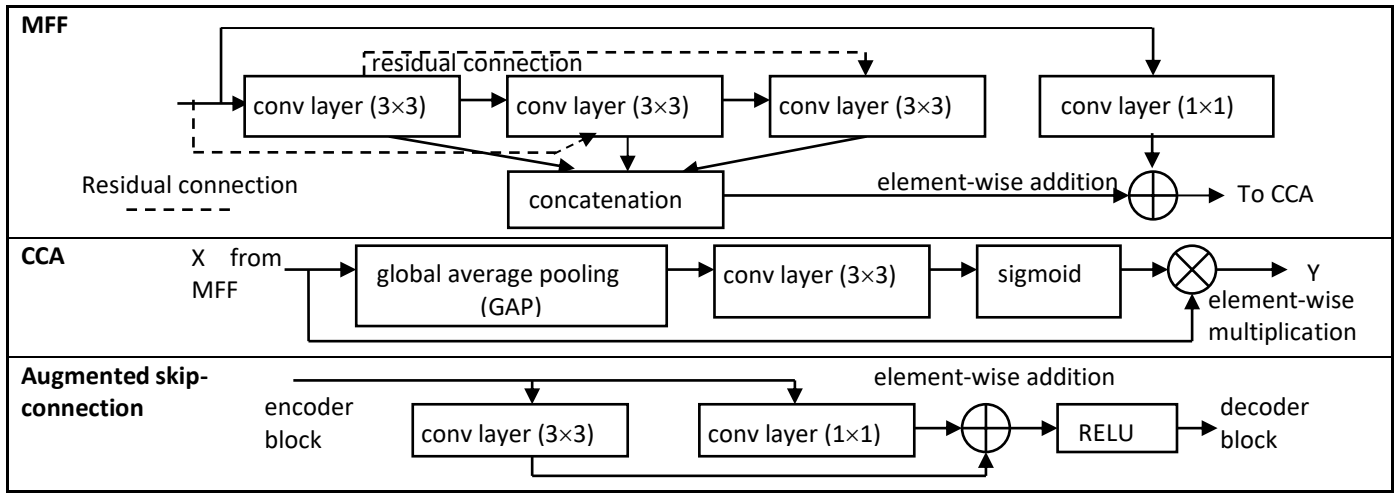

Fig. 2: A schematic of encoder-block components

MFF comprises two parts: (1) multiple convolution blocks for feature-extraction; (2) concatenating outputs from convolution blocks to keep diverse feature representations from various stages of extraction. MFF also uses residual connections within convolutional blocks. 1x1 convolutional kernel adjusts the weights of each feature-map (or the intensity of pixels for the first encoder-block), identifying important regions. The sum of features of all three convolution blocks is the same as the number of features in 1×1 convolution block; The vectors derived after concatenation and 1x1 convolution have the same dimension. Fused features from convolution-blocks are filtered using ReLU. The values in 1×1 kernel are adjusted using backpropagation to minimize the overall dice-loss. The two vectors are added element-wise.

CCA employs GAP followed by Conv2D and sigmoid activation function. GAP extracts global descriptors of each feature-map, capturing its overall characteristics. CCA uses a convolutional filter (conv2D) of kernel size 3×3. By applying convolutions, the descriptors representing channels, acquired after GAP, show cross-channel interactions.

The sigmoid function maps the values into the range [0, 1], which is interpreted as attention-weights for each window. The adjusted weights enhance relevant windows and suppress irrelevant ones. The output feature-map vector of CCA is calculated using Eq. (1).

$$Y = X \otimes (\sigma(\text{conv2D}(W \otimes \text{GAP}(X)))) \qquad (1)$$

The variable X denotes the input feature-map vector. The variable Y denotes the output feature-map vector. GAP (·) denotes the *global average pooling*. The capital letter W denotes the vector of convolutional kernel weights. The symbol σ (·) denotes the sigmoid function. The symbol ⊗ denotes element-wise multiplication of two vectors. Finally, the output of sigmoid function is combined with X, using element-wise multiplication, to obtain the final output Y.

### 4.2 Augmented Skip Connection

Residual links in the skip connection alleviate the problem of information loss at the corresponding decoder (see Fig. 2). This involves convolving the encoder output with both 3x3 and 1x1 kernels in parallel, allowing for the fusion of features from different abstraction levels. The resulting feature maps are element-wise summed and passed through ReLU activation function. This helps maintain the continuity of feature representations throughout the network, reducing the risk of information loss or degradation. During forward pass, fused features are compared with ground truth using Dice coefficient to calculate the dice loss. Compared to just using 1 x 1 kernel, combining 3 x 3 kernel with 1 x 1 kernel gave better accuracy at the cost of increased trainable parameters; The accuracy of renal tumor segmentation increased by 1% when both 3 X3 and 1x1 were combined.

## 5. Implementation

The software was developed using Python 3.9 programming language, using the TensorFlow framework. The model w as trained on a single processor in a processor-cluster, accelerated by an NVIDIA Tesla V100 GPU and 32 GB of RAM, su pported by CUDA 11.8.0, and cuDNN version 8801.

### 5.1 Dataset

We used publicly available KiTS19 Challenge database [21]. It contains 210 contrast-enhanced CT scans in NIFTI format, obtained from patients, who underwent partial or radical nephrectomy for kidney tumors. The slice thickness of the scans varied from 0.5 mm to 5.0 mm across different cases. Ground truth labels were manually annotated and confirmed by an expert clinician, using axial projections of the images [21]. The sample images for kidney and renal tumor, along with their masks, are illustrated in Fig. 3.

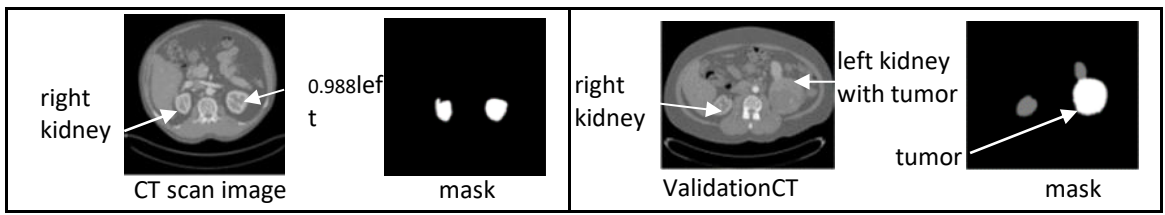

Fig. 3: Sample axial CT scan images and corresponding masks

## 5.2 Preprocessing

NIFTI images were converted into JPG format, resulting in 1,07,520 images. 16,336 images displaying anatomical organs were retained after filtering out black images. Retained images were resized to 256 pixels x 256 pixels using nearest-neighbor interpolation, preserving sharp edges and details crucial for medical imaging. Finally, pixel values were normalized by dividing each value by 255 to ensure they fell within the normalized range of [0, 1].

## 5.4 Evaluation Metrics

The model was evaluated using two metrics: Dice similarity coefficient (DSC) and Jaccard Index (JI). DSC is calculated using Eq. (2). The variable X shows the set of pixels in the ground truth; the variable Y shows the set of pixels in the predicted segment. Higher DSC shows higher overlap between the ground-truth and the predicted segment.

$$DSC(X, Y) = (2 \times |X \cap Y|) / (|X| + |Y|) \tag{2}$$

The Jaccard Index (JI), also known as the Intersection over Union (IoU), is computed using Eq. (3). X is the set of pixels in the ground truth; Y is the set of pixels in the predicted segment. Higher JI exhibits higher similarity between the ground-truth and the predicted segment.

$$JI(X, Y) = |X \cap Y| / |X \cup Y| \tag{3}$$

## 5.3 Training

The dataset was split into training (9801 samples), validation (3267 samples), and test sets (3268 samples). The model was trained for 50 epochs with early stopping on the validation loss plateau. *Adaptive Moment Estimation* (ADAM) optimizer was used for training with a batch size of 2. Dice loss (DL), calculated as *1 - DSC*, is a measure of dissimilarity between the predicted and ground-truth segmentations. Dice loss was applied for backpropagation to improve the training. Dice loss and DSC were tracked for the validation and testing to assess the accuracy achieved by the model.

## 6. Results and Discussion

Figure 4 illustrates the renal tumor segmentations derived using our model. Figure 5 shows the training and validation characteristics of our model for renal tumor segmentation. During training for renal tumor segmentation, DSC of 0.999 and dice loss of 0.001 was achieved after 50 epochs. The corresponding DSC and loss during validation (for 50 epochs) are 0.998 and .002, respectively. The average value of JI during validation of kidney segmentation was 0.95. During testing for renal tumor segmentation, the model attained DSC of 0.96 and JI of 0.91, respectively.

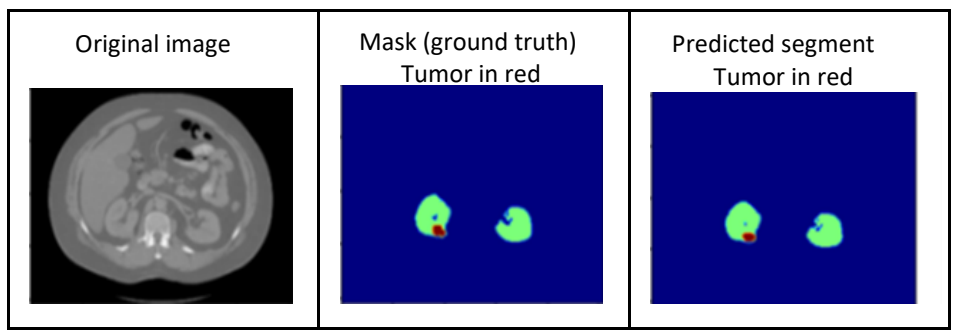

Fig. 4: Ground truth vs predicted masks for renal tumor segmentation.

As shown in Table 1, the model's accuracy outperformed existing leading models in segmenting renal tumors, due to its ability to capture subtle tumor-features, facilitated by the integration of MFF, CCA, and skip-connection enhancements.

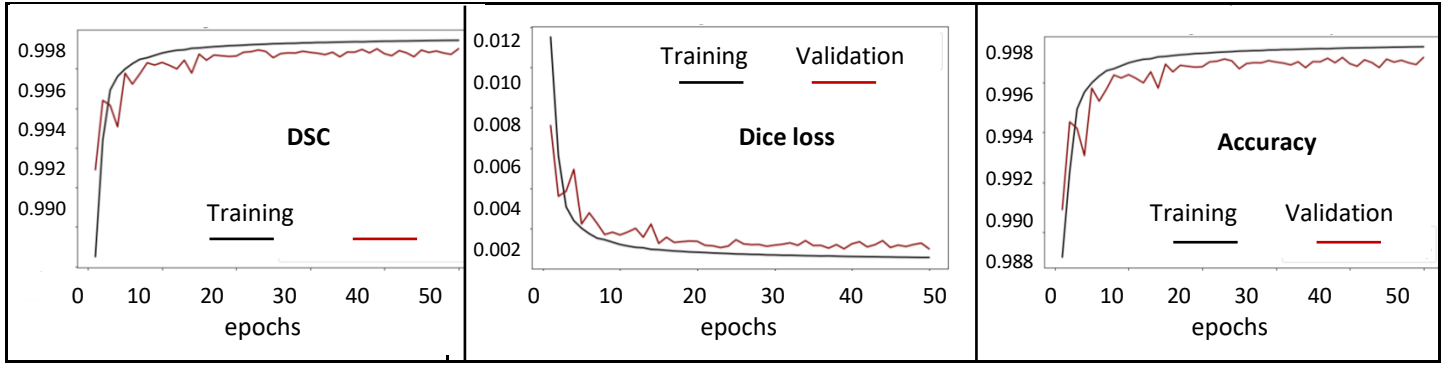

Fig. 5: DSC, dice loss, and accuracy of renal tumor segmentation during training and validation for multiple epochs

Table 1: Comparison of DSC in our model with other leading models for kidney and renal tumor segmentation

| model | segment. DSC kidney | segment. DSC tumor | model | segment. DSC kidney | segment. DSC tumor |
|---|---|---|---|---|---|
| [13] ensemble of two U-Nets | 0.95 | 0.61 | [17] residual + attention + U-Net | 0.96 | 0.77 |
| [14] AlexNet (residual connections) + U-Net | 0.96 | -- | [18] Swin transformer + UNet | 0.97 | 0.85 |
| [15] squeeze-and-excitation + residual U-Net | 0.92 | 0.54 | **Our approach** | **0.97** | **0.96** |
| [16] spatial + channel attention + U-Net | 0.94 | 0.68 | | | |

## 7. Conclusion and Future Work

Automated renal tumor segmentation is important for cost-effective early-stage diagnosis. We have proposed a U-Net variant for renal tumor segmentation with significantly enhanced accuracy over other leading models. The segmentation accuracy has been enhanced by integrating residual links between convolution layers and attention in the encoder-blocks and passing enriched information through the skip connections.

The model can be extended to segment tumors in other vital organs, such as liver. The current segmentation focuses on axial slices only. It lacks the information that can be derived from coronal and sagittal slices Our next step is to do multi-axial analysis by fusing the information from coronal and sagittal axis to further improve the accuracy. We are also working on the classification of tumors into benign or malignant by analyzing heterogeneity of tumors.

## Acknowledgements

We thank Ohio Supercomputing Center (OSC) for the use of supercomputing cluster to train our model through an educational grant.